\documentclass[review]{elsarticle}
\graphicspath{ {./figures/} }
\usepackage{float}
\usepackage{verbatim} 
\usepackage{multirow}
\usepackage{array}
\usepackage{booktabs}
\restylefloat{figure}
\floatstyle{plaintop} 
\restylefloat{table}
\usepackage[margin=0.80in]{geometry}

\usepackage{ccaption}
\usepackage{dcolumn}
\usepackage{multirow}
\usepackage[table]{xcolor}
\usepackage{wrapfig}
\usepackage{pifont}
\usepackage{float}
\usepackage{adjustbox}
\usepackage{wrapfig}
\usepackage{soul}
\usepackage{footnote}
\usepackage{times}
\usepackage{lipsum}
\usepackage{lineno}

\usepackage[T1]{fontenc}
\usepackage{textcomp}
\usepackage{graphicx}
\usepackage{graphics}
\usepackage{soul}

\usepackage{epstopdf}

\usepackage{caption}
\usepackage{subcaption}

\usepackage{amssymb,amstext,amsmath}
\usepackage{float}
\usepackage{booktabs}
\usepackage{array,ragged2e}
\usepackage[table]{xcolor}
\usepackage{multirow}
\usepackage{algorithm}
\usepackage{algpseudocode}
\usepackage{pifont}
\usepackage{enumerate}

\usepackage{hyperref}

\usepackage{natbib}
\biboptions{square,numbers,sort&compress}

\journal{Expert Systems with Applications}

\begin{document}
\begin{frontmatter}











\title{Unraveling Pedestrian Fatality Patterns: A Comparative Study with Explainable AI}

\author[label1]{Methusela Sulle \corref{cor1}}
\ead{msulle@scsu.edu}

\author[label1]{Judith Mwakalonge}
\ead{jmwakalo@scsu.edu}

\author[label2]{Gurcan Comert}
\ead{gcomert@ncat.edu}

\author[label1]{Saidi Siuhi}
\ead{ssiuhi@scsu.edu}

\author[label1]{Nana Kankam Gyimah}
\ead{ngyimah@scsu.edu}

\cortext[cor1]{Corresponding author.}
\address[label1]{Department of Engineering, South Carolina State University, Orangeburg, South Carolina, USA, 29117}
\address[label2]{Department of Computational Data Science and Engineering, North Carolina A\&T State University, Greensboro, North Carolina, US, 27411}

\begin{abstract}
Road fatalities pose significant public safety and health challenges worldwide, with pedestrians being particularly vulnerable in vehicle-pedestrian crashes due to disparities in physical and performance characteristics. This study employs explainable artificial intelligence (XAI) to identify key factors contributing to pedestrian fatalities across the five U.S. states with the highest crash rates (2018–2022). It compares them to the five states with the lowest fatality rates. Using data from the Fatality Analysis Reporting System (FARS), the study applies machine learning techniques—including Decision Trees, Gradient Boosting Trees, Random Forests, and XGBoost—to predict contributing factors to pedestrian fatalities. To address data imbalance, the Synthetic Minority Over-sampling Technique (SMOTE) is utilized, while SHapley Additive Explanations (SHAP) values enhance model interpretability. The results indicate that age, alcohol and drug use, location, and environmental conditions are significant predictors of pedestrian fatalities. The XGBoost model outperformed others, achieving a balanced accuracy of 98\%, accuracy of 90\%, precision of 92\%, recall of 90\%, and an F1 score of 91\%. Findings reveal that pedestrian fatalities are more common in mid-block locations and areas with poor visibility, with older adults and substance-impaired individuals at higher risk. These insights can inform policymakers and urban planners in implementing targeted safety measures, such as improved lighting, enhanced pedestrian infrastructure, and stricter traffic law enforcement, to reduce fatalities and improve public safety.

\end{abstract}


\end{frontmatter}

\section{Introduction}
Road-related pedestrian fatalities represent a significant public health and safety challenge, accounting for a substantial portion of road traffic deaths globally \cite{SCHWARTZ2022102896}. According to the World Health Organization (WHO), over 1.35 million lives are lost annually in road crashes, with an additional 20 to 50 million sustaining injuries \cite{hossain2024factors}. Vulnerable road users, including pedestrians, cyclists, and motorcyclists, constitute over 50\% of these fatalities \cite{usdot_vulnerable_road_users_2024}.
In the U.S., pedestrian fatalities increased by 53\% between 2009 and 2018, rising from 4,109 to 6,374 deaths \cite{tefft2021examining}, with a further 18\% increase from 2018 to 2022 \cite{nhtsa_fars}. Contributing factors include unsafe pedestrian behaviors, impaired driving, inadequate infrastructure, and poor visibility \cite{macioszek2023identification, campbell2004pedestrian}. Despite advancements in vehicle safety technologies, pedestrians remain disproportionately at risk due to their lack of physical protection \cite{khan2024advancing}. Addressing this issue requires a comprehensive understanding of contributing variables and the development of targeted interventions to mitigate pedestrian fatalities.
Despite efforts to address the identified factors, the persistent rise in pedestrian fatalities underscores significant gaps in our understanding and intervention strategies. While infrastructure improvements and public awareness campaigns have been widely implemented, their limited impact suggests that current approaches may overlook the complex interplay of dynamics contributing to these incidents \cite{CRAIGSCHECKMAN2024104891}. Existing studies often analyze data at national or regional levels, overlooking localized trends and high-risk zones that could provide actionable insights, moreover, many traditional methodologies fail to clearly understand the underlying causes, limiting their utility in developing targeted interventions. These challenges necessitate advanced analytical methods to identify high-risk areas and address the complexity of the factors driving pedestrian fatalities.

Addressing pedestrian fatalities requires overcoming significant gaps in localized risk assessment and multifactorial analysis. However, analyzing these issues presents several key challenges, which are outlined below:
\begin{enumerate}
    \item High variability in infrastructure, traffic patterns, and pedestrian behavior complicates the identification of high-incidence zones for pedestrian fatalities  \cite{su17020776}.
    \item The interplay of contributing factors road characteristics, vehicle attributes, and environmental conditions combined with data sparsity and inconsistencies make accurate analysis and explanation of these fatalities challenging  \cite{WANG2013264}.
    \item Variations in model performance caused by data distribution and prediction complexity highlight the need for an ensemble learning approach that combines the strengths of multiple algorithms to ensure accurate and reliable predictions of contributing factors  \cite{KHAN2024122778}.
    \item The lack of interpretability in conventional machine learning models hinders understanding and enhancing transparency and reduces practical applicability \cite{RASHEED2022106043}.
\end{enumerate}

Numerous studies have explored methods to reduce pedestrian fatalities, but critical gaps remain. Stigson et al. \cite{stigson2023reduce} demonstrated the safety benefits of vehicle and infrastructure modifications but focused primarily on high-income areas, limiting applicability to diverse U.S. contexts. Billah et al. \cite{billah2021analysis} analyzed spatial factors influencing pedestrian safety, providing insights into high-risk zones, but lacked predictive modeling for actionable approaches. Al-Ani et al. \cite{al2023predicting} employed deep learning models with high accuracy but failed to incorporate explainable Artificial Intelligence (AI)  or address data imbalances that bias predictions. These limitations underscore the need for a comprehensive framework integrating localized analysis, robust predictive models, and explainable AI to deliver interpretable and practical solutions for reducing pedestrian fatalities.

To address the challenges, this study applies a clustering algorithm to identify high-risk zones and conducts exploratory data analysis to examine the factors contributing to pedestrian fatalities. Furthermore, we introduce the Ensemble Learning Framework (ELF), which integrates predictive ensemble learning modeling with SHAP XAI to provide interpretable explanations and identify the key influences driving these fatalities. The key contributions of this work are as follows:
\begin{enumerate}
    \item Conducted exploratory data analysis to identify high-risk zones and address the variability in infrastructure, traffic patterns, and pedestrian behavior.
        
    \item Propose an ensemble learning model that combines the strengths of multiple algorithms to deliver robust and accurate predictions while overcoming the limitations of individual models.
    
    \item Integrate the ensemble learning model with SHAP (SHapley Additive Explanations), an XAI framework, to generate interpretable predictions, provide insights into feature importance, and enhance transparency in detecting pedestrian fatalities.
\end{enumerate}

The rest of this paper is organized as follows: Section $\text{II}$ presents related works on pedestrian fatality prediction. Section $\text{III}$ describes the proposed methodology. Section $\text{IV}$ describes the dataset, experiment setup, hyper-parameter settings, and evaluation metrics. Section $\text{V}$ presents the experimental results and discussion. Finally, the conclusion and future work are provided in Section $\text{VI}$.

\section{Related Works}
\subsection{Demographic and Behavioral Influences}
Various demographic and behavioral factors significantly influence pedestrian crashes, including age, gender, substance use, and distractions. Studies have identified substance impairment among pedestrians as a critical risk factor, with impairment significantly elevating the likelihood of fatal incidents\cite{dasilva2003pedestrian, nhtsa2016traffic, sun2019pedestrian}. For instance, Thomas et al. \cite{thomas2020epidemiology} showed that substance-related impairment accounted for a notable portion of pedestrian fatalities, underscoring the need for interventions targeting both behavioral and health-related aspects of road safety. Similarly, the increased use of smartphones among pedestrians has heightened the risk of accidents. Perry, \cite{perry2020smartphone} reported that texting or browsing while walking substantially increases the likelihood of accidents or near misses, emphasizing the urgency of addressing distracted behaviors.

\subsection{Environmental and Infrastructural Factors}
In Sweden, the work of Stigson et al. \cite{stigson2023reduce} provides an in-depth examination of the impact of both vehicle and infrastructure modifications on reducing pedestrian fatalities. Their findings suggest a promising potential for significant decreases in deaths by applying targeted safety measures such as autonomous emergency braking (AEB). However, they caution about delays due to slow adoption rates of new vehicle technologies. Low visibility conditions also play a vital part in increasing the pedestrian fatality rate in the US  \cite{ferenchak2021nighttime}. Roadway characteristics such as functional classification, number of lanes, and prevalent lighting conditions were also observed to affect pedestrian crash severity \cite{sun2019pedestrian}. Islam and Jones, \cite{islam2014pedestrian} found that pedestrian crash severity is influenced by dark lighting conditions and the number of lanes (two-lane roadways).

\subsection{Geographical and Comparative Studies}
In the United States, distinct regional studies provide insights tailored to local conditions. For instance, the analysis of pedestrian-vehicle crashes in San Antonio, Texas, by Billah \cite{billah2021analysis} highlights the critical role of spatial and environmental factors when designing safety interventions. The study suggests that location-specific measures, such as improving crosswalk visibility and adjusting traffic signal timings, could significantly enhance pedestrian safety, particularly in urban settings with high traffic volumes and complex interaction zones.

Similarly, Fox et al. \cite{fox2015spatiotemporal} utilized geocoded pedestrian mortality data from the Cali Injury Surveillance System (2008–2010) and census data to visualize and estimate pedestrian fatalities accurately. By mapping the spatiotemporal distribution of pedestrian mortality rates, the study identified high-priority areas for targeted prevention strategies. Key contributing factors related to geographical locations included traffic density, proximity to commercial zones, inadequate pedestrian infrastructure, and variations in land use patterns. While these geographical studies provide valuable insights into high-risk areas and contextual factors, they primarily focus on descriptive and comparative analyses. To develop proactive safety measures, it is essential to transition from understanding where incidents occur to predicting when and why they are likely to happen.

\subsection{Predictive and Analytical Modeling}
Recent advancements in machine learning (ML) have significantly improved the predictive capabilities for analyzing pedestrian crash data. Traditional statistical models, which often assume linear relationships between variables, face limitations in capturing complex interactions among factors \cite{tao2022advanced}. ML models like Decision Trees, Random Forests (RF), and XGBoost have shown superior performance in handling non-linear relationships \cite{das2020application, guo2021older}. For example, Prato utilized Kohonen neural networks to analyze fatal pedestrian crashes, identifying risk patterns such as nighttime crossings and male demographics \cite{prato2012mapping}. More recent studies have incorporated deep learning techniques, achieving higher accuracy and adaptability. For instance, Khan \cite{khan2024predicting} employed the Inception-v3 model combined with the Boruta algorithm to identify key contributing factors such as visibility and speed limits, addressing data imbalance through SMOTE. Similarly, Al-Ani et al. \cite{al2023predicting} used TabNet to predict pedestrian fatalities, highlighting its interpretive capabilities and high recall rates in identifying influential factors like roadway geometry and lighting conditions. Despite these advancements, existing studies rarely incorporate interpretability techniques like SHAP to explain model predictions, limiting their applicability for policymaking.

While prior research has provided valuable insights into demographic, environmental, and predictive modeling aspects of pedestrian fatalities, critical gaps remain. This limits the understanding of how state-level variations influence fatality rates and the identification of localized best practices. Additionally, while advanced ML techniques have improved predictive accuracy, the lack of interpretable frameworks hinders the translation of findings into actionable policies. This study addresses these limitations by focusing on the top and bottom five U.S. states in terms of pedestrian fatality rates, leveraging advanced ML techniques such as SMOTE for data balancing and SHAP for model interpretability. This comprehensive approach uncovers hidden patterns in the data and provides robust, actionable insights to guide the development of informed safety strategies.

\section{Proposed Methodology}
\begin{figure*}
    \centering
    \includegraphics[width=0.8\linewidth]{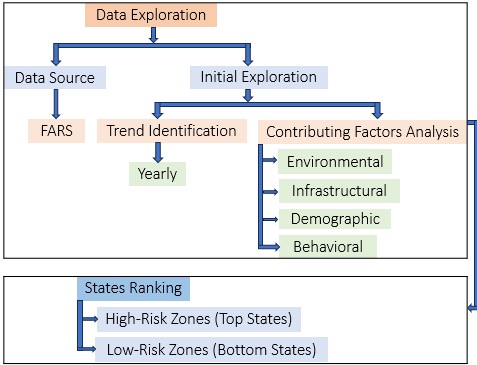}
    \caption{Identification of high-risk pedestrian fatality zones}
    \label{fig:High_risk_zones}
\end{figure*}

The proposed methodology consists of two distinct phases: identifying high-risk pedestrian fatality zones and the exploratory data analysis phase, as summarized in  Fig.~\ref{fig:High_risk_zones} the Ensemble Learning Framework (ELF) integrates predictive ensemble learning modeling with SHAP XAI to provide interpretable explanations and identify key factors driving these fatalities, as illustrated in Fig.~\ref{fig:proposed_framework}.

\subsection{Data Exploration and High-Risk Zone Identification}

\subsubsection{Data Exploration}
An initial dataset assessment to understand its structure, quality, and key characteristics to uncover trends, detect anomalies, and identify potential relationships between variables, laying the foundation for deeper analysis in Table \ref{fig:High_risk_zones}.

\subsubsection{High-Risk Zone Identification}
Pinpointing areas where pedestrian fatalities occur more frequently. This involves analyzing crash data for patterns like high-incidence locations like intersections, highways, or urban areas. Factors such as poor infrastructure (e.g., lack of crosswalks or sidewalks), high-speed zones, and lighting conditions are typically examined, Table \ref{fig:High_risk_zones}

\subsection{Comparative Analysis of Machine Learning Models}
\begin{figure*}
    \centering
    \includegraphics[width=0.8\linewidth]{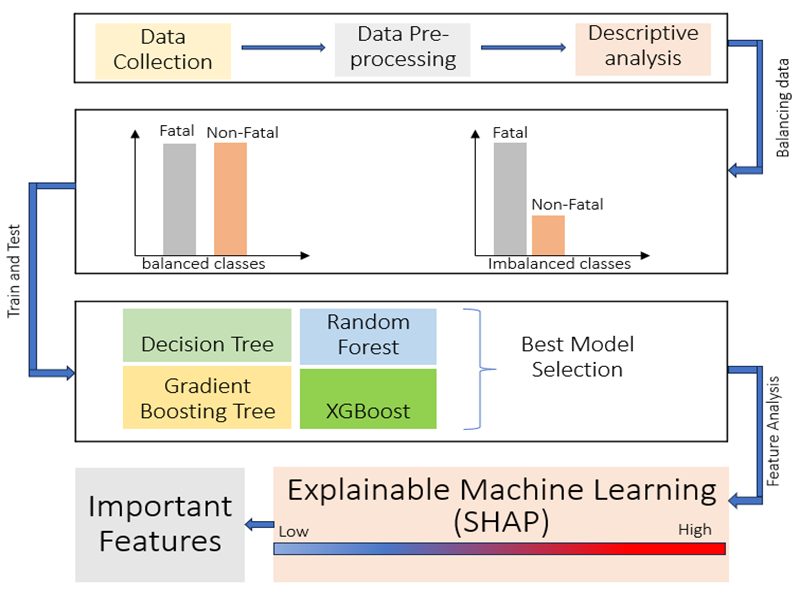}
    \caption{Feature analysis flow using Machine Learning methods}
    \label{fig:proposed_framework}
\end{figure*}

This study conducted a comprehensive comparative analysis of various machine learning (ML) models to determine the most effective approach for predicting pedestrian fatalities. Rather than focusing solely on the advantages of ensemble learning, we systematically compared the performance of individual and ensemble-based models. The primary objective was to evaluate and identify the best-performing model based on predictive accuracy and robustness. Specifically, the models analyzed included Decision Tree (DT), Gradient Boosted Trees (GBTs), Random Forest (RF), and Extreme Gradient Boosting (XGBoost) as illustrated in Fig.~\ref{fig:proposed_framework}, to evaluate model performance and identify the most effective approach for predicting pedestrian fatalities. Specifically, we employed Decision Tree (DT), Gradient Boosted Trees (GBTs), Random Forest (RF), and Extreme Gradient Boosting (XGBoost) models.

\subsubsection{Extreme Gradient Boosting (XGBoost)}
This is a scaleable and efficient implementation of gradient boosting designed for large-scale predictive modeling tasks. It incorporates advanced features such as regularization, missing data handling, and parallel computation, making it particularly suited for complex datasets. The model iteratively builds an ensemble of decision trees by minimizing a regularized objective function \cite{10.1145/2939672.2939785}, as defined in Eq.~\ref{eq:xgboost}.

\begin{align}
    \mathcal{L}(\theta) = \sum_{i=1}^{n} l(y_i, \hat{y}_i) + \sum_{k=1}^{K} \Omega(f_k),
    \label{eq:xgboost}
\end{align}

Where \( l(y_i, \hat{y}_i) \) represents the loss function, and \( \Omega(f_k) \) penalizes model complexity to mitigate overfitting.  

At each iteration, XGBoost updates predictions by adding a new tree that corrects residual errors, as formulated in Eq.~\ref{eq:residuals}:

\begin{equation}
    \hat{y}_i^{(t)} = \hat{y}_i^{(t-1)} + f_t(x_i),
    \label{eq:residuals}
\end{equation}

where \( f_t(x_i) \) represents the contribution of the newly added tree. The optimization process leverages first-order gradients \( g_i \) and second-order Hessians \( h_i \), ensuring both precision and computational efficiency.  

To prevent overfitting while maintaining performance, XGBoost employs a regularization term, as defined in Eq.~\ref{eq:xgboost-reg}:

\begin{align}
    \Omega(f_k) = \gamma T + \frac{1}{2} \lambda \|\omega\|^2,
    \label{eq:xgboost-reg}
\end{align}

where \( \gamma T \) penalizes the number of leaves \( T \), while \( \lambda \|\omega\|^2 \) regularizes the magnitude of leaf weights \( \omega \). These features, combined with parallel computation, make XGBoost a powerful and reliable tool for predictive modeling across diverse applications.

\subsubsection{Gradient Boosting Trees (GBTs)} 
This is powerful
and flexible implementation of gradient boosting is widely used for predictive modeling tasks such as regression and classification. GBTs build an ensemble of decision trees sequentially, with each tree correcting the residual errors of the previous ensemble. This iterative process minimizes a loss function through gradient descent, enabling GBTs to handle complex datasets effectively.
 \cite{rizkallah2025enhancing}.

The objective function of GBTs minimizes the loss function \( \mathcal{L} \), as defined in Eq.~\ref{eq:gbt-loss}:

\begin{align}
    \mathcal{L} = \sum_{i=1}^{n} l(y_i, \hat{y}_i),
    \label{eq:gbt-loss}
\end{align}

Where \( l(y_i, \hat{y}_i) \) represents the log-loss function, \( y_i \) is the actual value for sample \( i \), and \( \hat{y}_i \) is the predicted value.  

At each iteration, GBTs compute the negative gradient of the loss function to estimate residual errors, as given in Eq.~\ref{eq:residual-error}:

\begin{equation}
    r_i^{(m)} = -\frac{\partial \mathcal{L}}{\partial \hat{y}_i^{(m-1)}}
    \label{eq:residual-error}
\end{equation}

Where \( r_i^{(m)} \) represents the residual for sample \( i \) at iteration \( m \). These residuals guide the construction of new trees that refine the model’s predictions. The updated ensemble model is formulated as follows:

\begin{equation}
    F_m(x) = F_{m-1}(x) + \nu \cdot h_m(x),
    \label{eq:predictions}   
\end{equation}

where \( F_m(x) \) is the ensemble model after \( m \) iterations, \( h_m(x) \) denotes the newly added decision tree, and \( \nu \) is the learning rate, which controls the contribution of each tree to the final model.  
Regularization techniques, such as limiting tree depth and subsampling, prevent overfitting and enhance generalization by training each tree on a random data subset. This robustness and the ability to handle diverse loss functions and capture complex patterns make GBTs a reliable algorithm for predictive modeling across various applications.

\subsubsection{Random Forest (RF)}
This robust earning algorithm builds an ensemble of decision trees, each trained on a bootstrapped subset of the data it uses bootstrapping and feature sampling to ensure tree diversity, reducing overfitting and improving generalization. For classification, it minimizes classification error by reducing impurity, such as the Gini index \cite{SUN2024121549}. The final prediction is obtained via majority voting (for classification) or averaging (for regression), as defined in Eq.~\ref{eq:rf-aggregate}:

\begin{align}
    \hat{y} = \text{Aggregate}(T_1(x), T_2(x), \dots, T_n(x)),
    \label{eq:rf-aggregate}
\end{align}

where \( T_i(x) \) represents the \( i \)-th tree’s prediction.  

RF reduces impurity using the Gini index (Eq.~\ref{eq:rf-gini}):

\begin{align}
    \text{Gini Index} = 1 - \sum_{k=1}^{K} p_k^2,
    \label{eq:rf-gini}
\end{align}

where \( p_k \) denotes the proportion of samples in class \( k \) at a given node.  


\subsubsection{Decision Trees (DTs)}
This versatile predictive modeling technique partitions data into subsets by recursively splitting based on feature values, forming a hierarchical tree structure where internal nodes represent decisions and leaf nodes provide final predictions \cite{PACKWOOD2022100265}.  

The tree optimizes a splitting criterion to maximize subset purity. The Gini index metric, which measures the impurity of each node, is used, as defined in Eq.~\ref{eq:gini-index}:

\begin{align}
    \text{Gini Index} = 1 - \sum_{k=1}^{K} p_k^2,
    \label{eq:gini-index}
\end{align}

where \( p_k \) represents the proportion of samples in class \( k \), and \( K \) is the total number of classes. A lower Gini index indicates a purer node.    


\subsection{Post-hoc Model Interpretations}
While ensemble models provide high predictive accuracy, interpretability is crucial for transparency. Post-hoc analysis clarifies decision-making, ensuring trust in ML models \cite{ahmad2018interpretable, sagi2020explainable}.  

In this study, we use SHAP \cite{lundberg2017unified}, a perturbation-based method as shown in Fig. \ref{fig:proposed_framework}, to interpret predictions. SHAP models input features as players in a cooperative game, attributing contributions via Shapley values:

\begin{align}
h(z') = \varnothing_0 + \sum_{i=1}^{N} \varnothing_i z'_i,
\label{eq:linear-exploratory}
\end{align}

Where \( h(z') \) is the explanation model, \( z' \) represents input features, and \( \varnothing_i \) denotes feature attribution.

TreeSHAP, a variant for tree-based models, quantifies feature impact using:

\begin{align}
\varnothing_i = \sum_{K \subseteq M \setminus \{i\}} \frac{|K|!(N - |K| - 1)!}{N!} \left[g_x(K \cup \{i\}) - g_x(K)\right],
\label{eq:attribution}
\end{align}

where \( K \) is a feature subset, \( M \) is the full feature set, and \( g_x(K) \) is the expected model output.

TreeSHAP enhances model transparency, ensuring interpretability in high-stakes domains.

\section{Experimental Setting}
This section describes the dataset, the experiment setup, the hyper-parameter tuning, and the performance evaluation metrics of the network intrusion detection model.

\subsection{Dataset}
\label{subsection:Dataset}
The data used in this study was retrieved from the Fatality Analysis Reporting System (FARS) from the National Highway Traffic Safety Administration (NHTSA) website. This data set includes detailed information about fatal motor vehicle crashes across all U.S. States. For the specified period, 2018-2022, We obtained annual data files that contain a comprehensive set of variables such as crash, vehicle, driver, and pedestrian-specific details. The dataset includes 128 variables and 36,322 observations;  in this study, the data was narrowed down to 14 variables see Table~\ref{Variable Names}) and 16,569 rows to match the aim of the study; this includes Top states (16352, 14) and (217, 14) for Bottom states shape (rows and columns). The variables were selected on their relation to contributions to Pedestrian Fatalities, such as crash type, location of crash characteristics, environmental factors, vehicle type(s), and personal (or user) factors that influence the severity of an accident, such as impairments and distraction traits. In this work, we consider the accident types based on injury severity. According to our data, we categorized pedestrian injury severity into two categories, Fatal and Non-Fatal, with the following definitions,

\begin{itemize}
    \item \textbf{Fatal Crash}: A road crash resulting in at least one pedestrian fatality.
    \item \textbf{Non-Fatal Crash}: A road crash without pedestrian fatalities.  
\end{itemize}

\begin{table*}[h!]
    \centering
    \caption{Selected Variables from the FARS Dataset.}
    \footnotesize
\begin{tabular}{l l}
\hline
\textbf{SAS Variable Name} & \textbf{Data Element Description} \\
\hline
State & State \\
Wrk\_Zone & Work Zone \\
Mdrdstrd & Driver Distracted \\
Mnmdstrd & Non-Motorist Distracted \\
Nmimpair & Condition (Impairment) at Time of Crash—Non-Motorist \\
Drimpair & Condition (Impairment) at Time of Crash—Driver \\
Weather1 & Atmospheric Conditions \\
Lgt\_Cond & Light Condition \\
Rur\_Urb & Rural vs. Urban Land Use \\
Age & Age \\
Inj\_Sev & Injury Severity \\
Drinking & Police Reported Alcohol Involvement \\
Drugs & Police Reported Drug Involvement \\
Location & Location of the Crash \\
\hline
\end{tabular}
    \label{Variable Names}
\end{table*}

\begin{table*}[h!]
    \centering
    \caption{Descriptive Statistics of Bottom States on Both Fatal and Non-Fatal Crashes}
    \footnotesize
    \begin{tabular}{l l c c c c c c c}
        \hline
        \multirow{2}{*}{\textbf{State Name}} & \multirow{2}{*}{\textbf{Injury Severity}} & \multicolumn{5}{c}{\textbf{Year}} & \multirow{2}{*}{\textbf{Total}} & \multirow{2}{*}{\textbf{Percentage \%}} \\
        \cline{3-7}
        & & \textbf{2018} & \textbf{2019} & \textbf{2020} & \textbf{2021} & \textbf{2022} & & \\
        \hline
        California & Fatal accidents      & 978 & 1011 & 1013 & 1179 & 1158 & 5339 & 33\% \\
                   & Non-fatal accidents  &  55 &   58 &   51 &   72 &   54 &  290 &  2\% \\
        Florida    & Fatal accidents      & 706 &  714 &  695 &  819 &  773 & 3707 & 23\% \\
                   & Non-fatal accidents  &  46 &   37 &   44 &   48 &   67 &  242 &  1\% \\
        Georgia    & Fatal accidents      & 262 &  236 &  279 &  307 &  345 & 1429 &  9\% \\
                   & Non-fatal accidents  &  19 &    8 &    7 &   12 &   16 &   62 &  0\% \\
        New York   & Fatal accidents      & 268 &  274 &  229 &  293 &  303 & 1367 &  8\% \\
                   & Non-fatal accidents  &  24 &   19 &   29 &   24 &   18 &  114 &  1\% \\
        Texas      & Fatal accidents      & 616 &  649 &  688 &  817 &  797 & 3567 & 22\% \\
                   & Non-fatal accidents  &  36 &   46 &   63 &   41 &   49 &  235 &  1\% \\
        \hline
        \textbf{Total} &                   & 5028 & 5071 & 5118 & 5633 & 5602 & 16352 & 100\% \\
        \hline
    \end{tabular}
    \label{pre-survey-top}
\end{table*}

\begin{table*}[h!]
    \centering
    \caption{Descriptive Statistics of Bottom States on Both Fatal and Non-Fatal Crashes}
    \footnotesize
    \begin{tabular}{l l c c c c c c c}
        \hline
        \multirow{2}{*}{\textbf{State Name}} & \multirow{2}{*}{\textbf{Injury Severity}} & \multicolumn{5}{c}{\textbf{Year}} & \multirow{2}{*}{\textbf{Total}} & \multirow{2}{*}{\textbf{Percentage \%}} \\
        \cline{3-7}
        & & \textbf{2018} & \textbf{2019} & \textbf{2020} & \textbf{2021} & \textbf{2022} & & \\
        \hline
        South Dakota  & Fatal accidents      & 10 &  7 & 14 & 14 & 11 & 56 & 26\% \\
                      & Non-fatal accidents  &  0 &  1 &  0 &  1 &  1 &  3 &  1\% \\
        Rhode Island  & Fatal accidents      &  7 &  8 & 17 &  7 &  7 & 46 & 21\% \\
                      & Non-fatal accidents  &  1 &  0 &  0 &  0 &  0 &  1 &  0\% \\
        North Dakota  & Fatal accidents      &  6 &  5 &  8 & 10 &  6 & 35 & 16\% \\
                      & Non-fatal accidents  &  0 &  0 &  0 &  2 &  1 &  3 &  1\% \\
        Vermont       & Fatal accidents      &  6 &  3 &  8 &  8 &  6 & 31 & 14\% \\
                      & Non-fatal accidents  &  0 &  0 &  0 &  0 &  0 &  0 &  0\% \\
        Wyoming       & Fatal accidents      &  6 & 11 &  6 & 11 &  7 & 41 & 19\% \\
                      & Non-fatal accidents  &  0 &  0 &  0 &  0 &  1 &  1 &  0\% \\
        \hline
        \textbf{Total} &                     & 36 & 35 & 53 & 53 & 40 & 217 & 100\% \\
        \hline
    \end{tabular}
    \label{pre-survey-bottom}
\end{table*}

\subsection{Experimental Setup}
Before addressing the class imbalance, the model's performance showed that 65.22\% of the actual negative cases were incorrectly classified. In contrast, the positive class demonstrated higher prediction accuracy at 93.47\%. The imbalance is evident through the high misclassification rate of the minority class, which significantly affects the model's overall performance, particularly in predicting minority class instances.

To address the class imbalance, this study employs sampling methods, specifically SMOTE for Nominal and Continuous features (SMOTENC), to improve model performance. SMOTENC generates synthetic instances based on k-nearest neighbors, balancing the dataset where fatal cases constitute 99.30\% for the negative class and 98.19\% for the positive class, reflecting enhanced model performance and reduced misclassification of minority class instances as seen in Fig. \ref{fig.Confusion Matrix}.

\subsection{Hyper-parameter Tuning}
The hyperparameters of the classifiers used for pedestrian fatalities prediction were determined through a grid search and 5-fold cross-validation. The cross-validation was performed on 80\% of the dataset allocated for training, while the remaining 20\% was reserved for testing. The hyperparameters identified through this process were subsequently used for model evaluation without further modification. Table \ref{table:hyperparameter setting}.

\begin{table}[h!]
\caption{Classifier hyperparameter settings.}  
\centering 
\footnotesize
{\begin{tabular}{c c c} 
\hline
Classifier &Hyper-parameter &Values\\ [0.3ex]
\hline
 &learning Rate & $0.3$\\
XGBoost & $max_{-}depth$ & $20$ \\
 &$n_{-}estimators$ & $100$\\
 &$eval_{-}metric$ & mlogloss\\
\hline
&C &0.1\\
GBT &Gamma &1 \\
  &Kernel &Linear\\
\hline
&C &0.1\\
Random Forest &Gamma &1 \\
  &Kernel &Linear\\
\hline
 &Criterion &Gini \\
 &$max_{-}depth$ &$9$ \\
DT &$max_{-}features$ &sqrt\\
 &$min_{-}samples_{-}leaf$ &$1$ \\
 &$min_{}-samples_{-}split$ &$2$\\
\hline 
\end{tabular}}
\label{table:hyperparameter setting}
\end{table}

\subsection{Evaluation Metrics}
\label{table:evaluation metrics}
The evaluation employed key performance metrics: accuracy, precision, recall, and F1 score, as defined in Eq. \ref{eq:accuracy} through Eq. \ref{eq:F1-score}, to assess the detection models’ performance. Additionally, the confusion matrix provides a visual representation of algorithm performance, expressed in terms of True Positives (TP), True Negatives (TN), False Positives (FP), and False Negatives (FN). These metrics comprehensively evaluate model effectiveness and reliability in predicting pedestrian fatalities.

\begin{equation}
    \text{Accuracy} = \frac{TP + TN}{TP + TN + FP + FN}
    \label{eq:accuracy}
\end{equation}

\begin{equation}
    \text{Precision} = \frac{TP}{TP + FP}
    \label{eq:precision}
\end{equation}

\begin{equation}
    \text{Recall} = \frac{TP}{TP + FN}
    \label{eq:recall}
\end{equation}

\begin{equation}
    \text{F1 Score} = \frac{2 \times \text{Precision} \times \text{Recall}}{\text{Precision} + \text{Recall}}
    \label{eq:F1-score}
\end{equation}

\hspace{1cm}
\section{Results}
\subsection{Model Selection}
Four classification models (Decision Tree (DT), Gradient Boosting Tree (GBT), Random Forest (RF), and XGBoost (XGB)) were used to predict pedestrian accidents (fatal and non-fatal) in both Top and Bottom identified states Table \ref{Models Performance}. Each model can show important features during training and demonstrate varying levels of accuracy, precision, recall, and F1 scores. In the case of imbalanced data, we introduced the confusion matrix Fig. \ref{fig.Confusion Matrix}, which gives balanced accuracy.

\begin{table*}[h!]
    \centering
    \caption{Results of classification ML models for Pedestrian Fatal and Non-Fatal crash severity prediction in Top and Bottom 5 states.
}
    \footnotesize

\begin{tabular}{l l l l l l}
 & \textbf{Balanced Accuracy} & \textbf{Accuracy} & \textbf{Precision} & \textbf{Recall} & \textbf{F1-score} \\
\hline
& & & & & \\
 XGBoost & 0.989952 & 0.906178 & 0.921224 & 0.906178 &0.913222 \\
Random Forest & 0.989952 & 0.898069 & 0.919704 & 0.898069 & 0.908064 \\
Gradient Boosting Tree & 0.963088 & 0.89305 & 0.925909 & 0.89305 & 0.907494 \\
Decision Tree & 0.989952 & 0.872973 & 0.916613 & 0.872973 & 0.892569 \\

\end{tabular}
\label{Models Performance}
\end{table*}

From the results, the XGBoost model emerged as the best in overall model performance with the highest Balanced accuracy and Accuracy, closely followed by the Random Forest. Also, the Gradient Boosting Tree showed robust performance in predicting pedestrian accidents, Decision Tree, while still effective, scored the lowest accuracy and F1 score among the models evaluated.

\subsection{XGBoost model}
The XGBoost model was selected as the primary model in our study due to its superior and outstanding performance metrics among other models tested in both Top and Bottom identified states, making it the most reliable model for predicting fatal and non-fatal pedestrian accidents. 

\begin{figure}[H]
\centering
\begin{subfigure}{0.5\columnwidth}
\centering
  \includegraphics[width=0.9\textwidth]{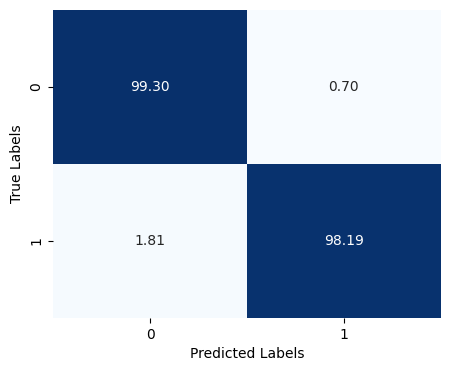}
\caption{Confusion Matrix.}
  \label{fig.Confusion Matrix}
\end{subfigure}%
\begin{subfigure}{0.5\columnwidth}
 \centering
\includegraphics[width=0.85\textwidth]{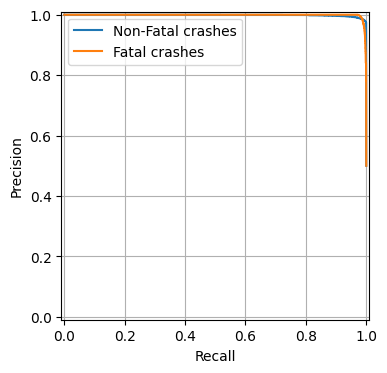}
  \caption{Precision-Recall Curve.}
  \label{fig.precision recall}
\end{subfigure}
\begin{subfigure}{0.5\columnwidth}
 \centering
\includegraphics[width=0.9\textwidth]{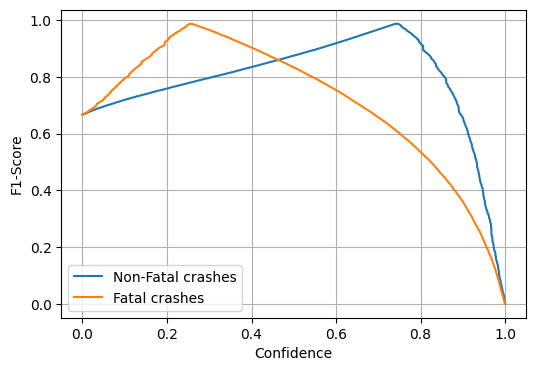}
  \caption{F1 Confidence curve.}
  \label{fir.F1}
\end{subfigure}
\caption{}
\end{figure}

Fig. \ref{fig.Confusion Matrix}: Shows the confusion matrix for Actual and Predicted Pedestrian Fatalities. The matrix attained TP of (98.30\%) and TN of (99.19\%), respectively, indicating that the model was highly accurate in predicting both classes. Also, the low percentages of FP (0.70\%) and FN  (1.81\%) demonstrate the model's reliability and precision.

The confusion matrix confirmed that the XGBoost model distinguishes fatal and non-fatal pedestrian accidents. The model’s ability to minimize the FN is critical in safety-critical applications, where accurately predicting fatal accidents can lead to more targeted and effective interventions.

Fig. \ref{fig.precision recall}: The precision-recall curves (PR curves). The curves show how the precision values vary with the increase in recall values during training \cite{ruseruka2023pavement}.  The curve plots precision (the ratio of true positive predictions to the total predicted positives) against recall (the ratio of true positive predictions to the total actual positives) for different threshold values.

Fig. \ref{fir.F1}: The curve indicates the balanced performance of precision and recall. Both curves, non-fatal and fatal crashes, exhibited a peak score of 99\%, indicating excellent model performance and an optimal confidence level.

Both the Precision-Recall curve and the F1 confidence curve in Fig. \ref{fig.precision recall} and Fig. \ref{fir.F1} highlight the effectiveness and robustness of the XGBoost model in predicting the severity of pedestrian accidents in both the Top and Bottom 5 States. These findings, therefore, underscore the model's capability in prediction. Using SMOTE to balance the dataset further enhances the model’s performance, ensuring accurate predictions across both non-fatal and fatal crash classes.

\subsection{Feature analysis using explainable ML.}
To understand the impact of contributing factors on pedestrian fatalities in identified states, we performed feature analysis from the model using explainable ML (SHAP) to uncover their relationship with the target variable.

\begin{figure}[H]
\centering
\begin{subfigure}{0.5\columnwidth}
\centering
  \includegraphics[width=0.9\textwidth]{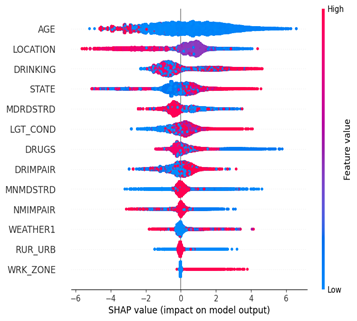}
\caption{Top 5 State's SHAP features importance.}
  \label{fig:shap top5 local}
\end{subfigure}%
\begin{subfigure}{0.5\columnwidth}
 \centering
\includegraphics[width=0.9\textwidth]{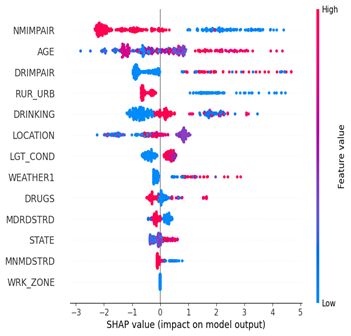}
  \caption{Bottom 5 State's SHAP features importance.}
  \label{fig:shap bottom5 local}
\end{subfigure}
\begin{subfigure}{0.5\columnwidth}
 \centering
\includegraphics[width=0.9\textwidth]{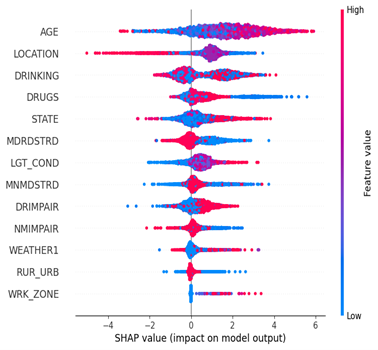}
  \caption{Combined State's local SHAP features importance.}
  \label{fig:combined local}
\end{subfigure}
\caption{}
\end{figure}

Fig. \ref{fig:shap top5 local} is the model SHAP features impact on pedestrian fatalities in the Top 5 states, likewise for the Bottom 5 states Fig. \ref{fig:shap bottom5 local}. Age is the most impactful predictor with higher values increasing the risk, whereas in the bottom states, the impact seen is mixed. Non-motorist impairment (NMIMPAIR) is the most significant predictor in the bottom states but less impactful in the top states. Drinking and drug use remain strong predictors in both sets of states, though their impact is more pronounced in the bottom states. Work zone (WRK\_ZONE) shows minimal impact in both sets, but it is slightly more negligible in the bottom states.

\subsection{Combined Dataset on Both Bottom and Top States}
We combined data from the top and bottom five U.S. states with the highest and lowest pedestrian fatalities to analyze and compare the contributing factors to pedestrian fatalities. This approach enabled us to identify the key determinant features and the differences between the states with high and low fatality rates  Fig. \ref{fig:combined local}, providing a holistic view of pedestrian safety across different regions.

The most influencing features, such as AGE, DRINKING, DRUG USE, LOCATION, and STATE, are discussed in the following section to understand their impact on pedestrian fatalities in both sets of the dataset.

\hspace{1cm}
\section{Discussion and Recommendation}
The main aim of this study was to predict contributing factors for pedestrian fatalities in the top five U.S. states with the highest records of such incidents from 2018 to 2022 and compare them to the bottom five states with the lowest rates. Utilizing data from the Fatality Analysis Reporting System (FARS). According to this study, machine-learning models are crucial in managing complex and substantial datasets. These algorithms offer an automated means of uncovering patterns and relationships within data, a task that can be challenging or infeasible when attempted manually \cite{elalouf2023developing}.
 Over the past decade, various studies have consistently favored ensemble models for multivariable pedestrian fatalities model development. This study utilized an ensemble method ((Decision Tree (DT), Gradient Boosting Tree (GBT), Random Forest (RF), and XGBoost (XGB)). This trend is found in studies like  \cite{ahmed2023study}, \cite{aboulola2024improving}, \cite{al2023predicting} and \cite{shanshal2020prediction}. These models showed the effectiveness of ensemble tree methods in pedestrian fatalities prediction, where in our study, XGBoost outperformed other models, achieving an accuracy of 90\% and a balanced accuracy of 98\%, making it consistent with the previous studies that used the classification models. While utilizing SHAP analysis to provide insight into feature importance, our study stood out for integrating the Synthetic Minority Oversampling Technique (SMOTE) to handle the imbalanced dataset, which has been investigated less in similar previous works.  Al-Ani et al. \cite{al2023predicting} focused on feature importance without full consideration of the imbalance issue, probably affecting the models' performance. Also, while in previous works, the model MobileNet showed an accuracy of 98.17\% \cite{aboulola2024improving}, by using ensemble models, the approach was more balanced with precision equal to 92\% and recall equal to 90\%-something important when we want to minimize false negatives in safety-critical applications.

In this study, we identified the following features as they have high contribution traits to pedestrian fatalities, Age of pedestrians has emerged as a significant predictor across all regions, and both young and elderly adults identified as high-risk Fig. \ref{fig:combined local}. Young pedestrians often exhibit impulsive behavior, such as running into the street, and may not fully understand traffic signals \cite{xiao2021study}. Older adults may have slower reaction time, impaired vision and hearing, and reduced mobility, making it challenging to avoid oncoming vehicles \cite{tournier2016review}. In Vermont, more than half of all pedestrians killed in motor vehicle crashes were over the age of 60 \cite{ghsa2022pedestrian}.  Location of the crashes (i.e., intersection or not at the intersection), The crash location, and the type of traffic control also play a significant role in pedestrian crashes \cite{ahmed2023study}. Most pedestrian fatalities occur at locations between intersections (i.e., midblock), where vehicles are more likely to drive straight and faster\cite{MOHAMMED2021584}. Drinking (fatalities caused by alcohol). The SHAP analysis highlights drinking as a critical predictor of fatal pedestrian crashes, particularly in states with higher rates of alcohol consumption. \cite{lasota2020alcohol} also highlighted that Alcohol is a significant risk factor for road accidents involving pedestrians as unprotected road users. Drugs (fatalities associated with drug substances): It is identified that Drug use, similar to alcohol, significantly impacts the likelihood of pedestrian fatalities. Drivers and pedestrians may develop a lack of coordination and awareness due to being under the influence of alcohol. NHTSA, \cite{nhtsa2022alcohol} indicated that drug prevalence among drivers and other road users, such as pedestrians and bicyclists, leads to serious or fatal injuries in crashes in the United States. State itself as a contributing factor to pedestrian fatalities; the top five states with high pedestrian fatality rates and the bottom five states with low fatality rates offer critical insights for safety interventions. States like California, Florida, and Texas experience higher fatalities. California \cite{heinrichlaw2024pedestrian}, reported that many cities have old and outdated infrastructure and increased pedestrian fatalities. As such, pedestrians must often cross the road unsafely. Brent C. and Miller \cite{miller2024florida} discuss Florida’s ranking as one of the deadliest states for pedestrians as it continues to experience a rise in pedestrian fatalities. Contributing factors are Poor Road designs, speeding drivers, distracted driving, and Lack of enforcement of existing laws, including speeding laws, distracted driving regulations, and rules requiring drivers to stop for pedestrians walking in a crosswalk. Likewise, the ranked 3rd State in fatal pedestrian accidents Texas  \cite{smithhassler2024texas}.At the same time, states like South and North Dakota, Wyoming, Rhode Island, and Vermont face fewer incidents. The use of seat belts by South Dakotans and community efforts such as law enforcement, schools, media, and safety organizations played a role in reducing the incidence of impaired driving and improving the safety of roadways to ensure those who are injured in traffic crashes receive a quick response and high-quality treatment \cite{sddps2023annual}. Vision Zero initiative in North Dakota, which requires that drivers and pedestrians have a shared responsibility to follow the rules of the road and act safely and responsibly, has contributed to the State’s lowest Pedestrian fatality record \cite{visionzero2022responsibility}. Broadly, lower traffic volumes and rural settings limit the traffic-pedestrian interactions in the bottom five states, which is also the reason for lower pedestrian fatality records. Driver and Non-motorist distractions were also shown to have an impact on pedestrian fatalities in all regions; Distracted driving is a significant factor in many pedestrian accidents. Drivers texting, talking on the phone, or eating while driving are more likely to miss pedestrians on the road \cite{msb2024pedestrian}. The use of digital devices among pedestrians has increased the risk of accidents. The article by Susan Perry \cite{sun2019pedestrian} reported that people use their smartphones to text or browse. At the same time, they walk considerably more risk of having an accident or a “near miss” than pedestrians who talk on their phones—further explained that it doesn’t mean that talking on a phone while walking isn’t risky. It’s just that the risk does not appear to be as great as that associated with texting. Light conditions (during the day or night) during the crash, whether during the day or night, significantly impact pedestrian safety. Nighttime accidents have higher SHAP values, indicating a greater likelihood of severe outcomes due to poor visibility. Research has found that pedestrians are at higher risk of a collision in the dark and that the severity of nighttime collisions is worse than that of daytime collisions, with nighttime pedestrian collisions at intersections having an  83\% higher chance of being fatal without street lighting and a 54\% higher chance of being fatal even with street lighting \cite{ferenchak2021nighttime}. Driver and non-motorist impairment, the condition of both drivers and pedestrians at the time of the crash, including impairments from alcohol, drugs, or fatigue, is crucial. SHAP values have shown the correlation between these two features and pedestrian fatalities. Weather conditions, such as rain, snow, or fog, affect visibility and road conditions, impacting pedestrian safety. Rural and Urban, the location of crashes in rural versus urban areas has different implications for pedestrian safety. NHTSA, \cite{nhtsa2021traffic} reported that more pedestrian fatalities occurred in urban areas (84\%) than rural areas (16\%) in 2021. Work zones often have confusing signage and reduced visibility, increasing the risk of accidents. While the model minimizes overall feature performance, they are still notable in certain areas with ongoing construction projects.

By employing the XGBoost model in our study, we have gained a comprehensive understanding of various factors and their impact on pedestrian safety, which allows for the development of targeted interventions and policies to reduce fatalities. This paper recommends that, in the coming period of the transportation era, safer strategies to minimize pedestrian fatalities and injury severity need to be specified, implemented, and enabled to monitor future performance. Strengthening the enforcement of traffic laws, particularly those related to high speeds, driving under the influence of alcohol and drugs, and distracted walking, is essential. Upgrading the pedestrian infrastructure, especially at unsignalized intersections and mid-block areas, is necessary. This includes the installation of better lighting and signals. The use of Bluetooth devices to alert the area pedestrian who engages themselves in digital devices that they are close to the crosswalk and the installation of ground lights at the crosswalk to remind distracted pedestrians about their whereabouts \cite{larue2020pedestrians} and \cite{schwebel2020using}. Sectors such as healthcare and law enforcement could use this information to develop comprehensive public awareness campaigns to educate the community about the importance of using designated crossing areas and the risk of distracted walking and driving. Department of Transportation (DOTs) should also prioritize continuously monitoring fatalities trends, especially in the top five states, to identify new high-risk areas and other contributor factors and allocate the resources to implement timely interventions effectively.

\section{Conclusion and Future Studies}
Road-related pedestrian fatalities are a public health concern in the United States. Utilizing data from the Fatality Analysis Reporting System (FARS), this study employed an ensemble ML approach to analyze the contributing factors to the crashes. SMONTEC and SHAP techniques were used to improve the dataset and explain the selected model. The results indicated that XGBOOST emerged as the best-performing model when the balanced dataset was used. It achieved a balanced accuracy of 98\%, accuracy (90\%), precision (92\%), recall (90\%), and score (91\%). The key contributing factors identified were age, alcohol and drug use, location, state, distractions and impairments of divers and pedestrians, lighting conditions, and environmental factors such as Weather and rural versus urban settings. Considering the policy recommendations, strengthening traffic law enforcement, upgrading pedestrian infrastructure, especially at unsignalized intersections, educating the community about the risks of distracted walking, and investing in modern technologies like vehicle-pedestrian detection systems. Future work should also continue to build on these insights. The findings of this work will provide valuable insight to transportation policymakers and urban planners in enabling targeted interventions to reduce road-related pedestrian fatalities and enhance public health and safety.

In Future works, researchers should focus on addressing the limitations identified in this study. Specifically, efforts should be made to balance the dataset by incorporating additional relevant features to improve model performance in predicting non-fatal crashes. Expanding the geographical scope beyond the Top and Bottom states and integrating GIS analysis can provide a more comprehensive understanding of location-specific factors influencing pedestrian fatalities. Additionally, future studies should explore other potentially significant variables, such as socioeconomic status, and conduct time-stability analyses to account for temporal variations in population distribution, road networks, and travel behavior.

\section{Credit authorship contribution statement}
\noindent \textbf{Methusela Sulle}: Conceptualization, Writing, and Data analysis. \textbf{Judith Mwakalonge}: Writing, data analysis, methodology, and conceptualization. \textbf{Gurcan Comert}:  Writing, data analysis, methodology, and conceptualization. \textbf{Saidi Siuhi}: Writing, data analysis, methodology, and conceptualization. \textbf{Nana Kankam Gyimah}: Writing, methodology, and conceptualization.

\section{Declaration of competing interest}
The authors declare that no competing financial interests or personal relationships could have influenced the work reported in this paper.

\section{Data availability statement}
This study used a publicly available dataset, and the information is included in this article.

\section{Acknowledgement}
The U.S. Department of Education supported this research through Grant No. P382G320015, administered by the Transportation Program at South Carolina State University (SCSU). The research was partly funded by the U.S. Department of Energy Minority Serving Institutions Partnership Program (MSIPP), TOA 0000525174 CN1, MSEIP 2 P20A210048, and NSF Grants Nos. 2131080, 2200457, 2234920, and 2305470.

\bibliography{sample.bib}

\end{document}